\documentclass[11pt]{article}

\usepackage[]{acl}

\usepackage{times}
\usepackage{latexsym}
\usepackage{enumitem}
\usepackage[T1]{fontenc}

\usepackage[utf8]{inputenc}

\usepackage{microtype}

\usepackage{inconsolata}

\usepackage{pifont}
\usepackage{graphicx}
\usepackage{subcaption}
\usepackage{amsmath}
\usepackage{amssymb}
\usepackage{multirow}
\usepackage{textcomp}
\usepackage{tikz}
\usepackage{booktabs} 
\usepackage{bm}
\usepackage{mathtools}
\usepackage{arydshln}
\usepackage{xcolor}
\usepackage{hyperref}

\title{Revisiting Catastrophic Forgetting in Large Language Model Tuning}

\author{Hongyu Li \\
  Wuhan University\\
  \texttt{hongyuli@whu.edu.cn} \\\And
  Liang Ding\thanks{Corresponding author.} \\
  The University of Sydney \\
  \texttt{liangding.liam@gmail.com} \\\AND
  Meng Fang \\
  University of Liverpool \\
  \texttt{mfang@liverpool.ac.uk} \\\And
  Dacheng Tao \\
  Nanyang Technological University \\
  \texttt{dacheng.tao@ntu.edu.sg}}

\begin{document}
\maketitle
\begin{abstract}
Catastrophic Forgetting (CF) means models forgetting previously acquired knowledge when learning new data. It compromises the effectiveness of large language models (LLMs) during fine-tuning, yet the underlying causes have not been thoroughly investigated. This paper takes the first step to reveal the direct link between the flatness of the model loss landscape and the extent of CF in the field of LLMs. Based on this, we introduce the sharpness-aware minimization to mitigate CF by flattening the loss landscape. Experiments on three widely-used fine-tuning datasets, spanning different model scales, demonstrate the effectiveness of our method in alleviating CF. Analyses show that we nicely complement the existing anti-forgetting strategies, further enhancing the resistance of LLMs to CF. 

\end{abstract}

\section{Introduction}

Instruction fine-tuning is key to improving the capabilities and controllability of large language models (LLMs)~\cite{instruction_tuning:2, instruction_tuning:3}, which have already demonstrated strong performance in various tasks~\cite{zhong2023chat,Peng2023ChatGPT4MT,Lu2023EAPrompt,ren2024healthcare}. One major obstacle to tuning LLMs is catastrophic forgetting (CF,~\citealp[]{kirkpatrick2017overcoming}), which means LLMs forget prior knowledge when learning new data.
Recent works have provided substantial evidence confirming the negative impact of CF on LLMs, e.g., \citet{bi2024deepseek} empirically show that the fine-tuned model is even worse than their foundation counterpart on several tasks, and \citet{agent} reveal the dropped general performance of LLMs after developing their agent capabilities.

Understanding the effectiveness of LLMs during fine-tuning is important for downstream tasks, however, the underlying causes of CF remain largely unexplored. There are two types of work addressing CF, from data and model perspectives, respectively. \citet{Rehearsal} propose continual learning with rehearsal in instruction tuning, though its effectiveness varies with task selection. \citet{Wise-ft} indicate that both continual learning and weight averaging (Wise-FT) effectively preserve generality.

While the above techniques are somewhat successful, they require expensive extra data-constructing and training costs, and are even sometimes impractical in LLMs, because a) the data cards of many pretrained models are unclear~\cite{shi2023detecting}, making rehearsal unfeasible, and b) anti-forgetting training brings an unstable and expensive training process~\cite{datta2023measuring}.

In this work, we turn to finding a cheap, stable, and orthogonal solution to alleviate the CF in tuning LLMs. In particular, we \ding{182} reveal the high correlation between the extent of CF and the flatness of the loss landscape (LLS,~\citealp[]{flatness_lls:1, flatness_lls:2}), \ding{183} mitigate the CF in LLMs by flattening the LLS from optimization perspective, \ding{184} show the complementarity between our method and existing anti-forgetting works.
Specifically, we designed three probing analyses to achieve \ding{182}, and found that a flatter LLS could reduce the severity of CF. Based on our observation, we introduced ``Sharpness-Aware Minimization'' (SAM,~\citealp[]{SAM}) to flatten the model LLS to approach \ding{183}. For \ding{184}, we found that our introduced optimizer nicely complements a series of anti-forgetting methods, including rehearsal~\cite{Rehearsal} and Wise-FT~\cite{Wise-ft}. 
Our \textbf{contributions} are as follows:

\begin{figure*}[ht]
    \centering
    \begin{subfigure}[b]{0.32\textwidth}
        \includegraphics[width=\textwidth]{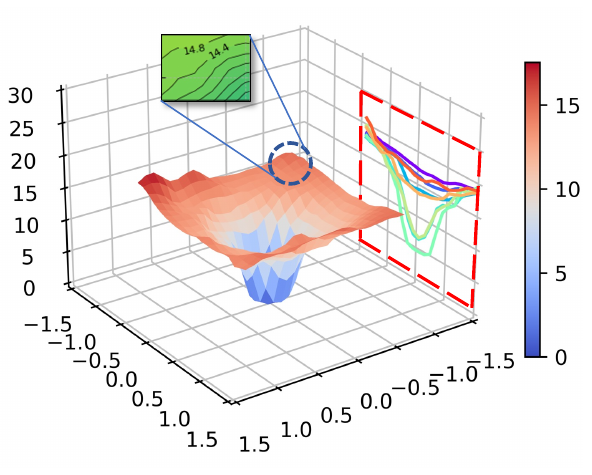}
        \caption{Alpaca.}
        \label{fig1:img1}
    \end{subfigure}
    \begin{subfigure}[b]{0.32\textwidth}
        \includegraphics[width=\textwidth]{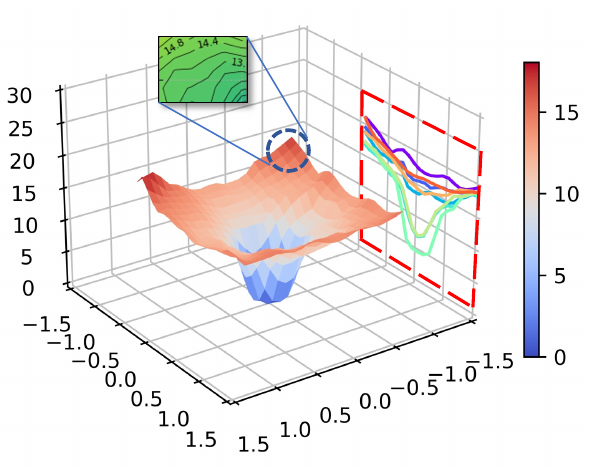}
        \caption{Alpaca $\rightarrow$ Open-Platypus}
        \label{fig1:img2}
    \end{subfigure}
    \begin{subfigure}[b]{0.32\textwidth}
        \includegraphics[width=\textwidth]{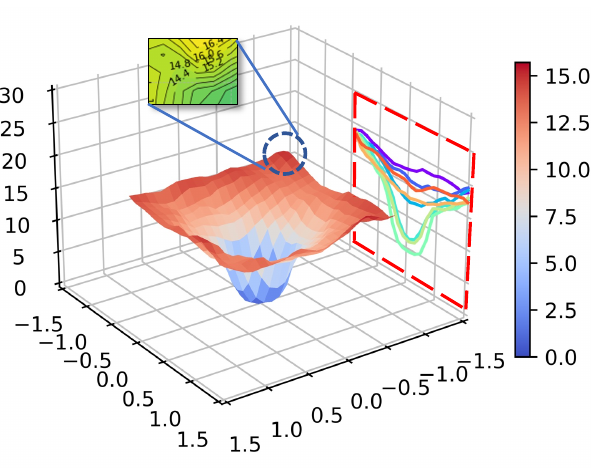}
        \caption{Alpaca $\rightarrow$ Auto-Wiki}
        \label{fig1:img3}
    \end{subfigure}
    \caption{\textbf{Visualization of different models' loss landscapes} (LLS) with contour lines. We can see that with the data/task gaps of (a), (b), (c) gradually increase, the disturbance of their contours becomes more obvious.}
    \label{fig:three_images}
\end{figure*}

\begin{itemize}
	\item To the best of our knowledge, we are the first to empirically reveal a direct correlation between the flatness of the model LLS and CF.
	\item We present the first work that mitigates CF of LLMs from the perspective of optimization.
        \item The proposed method can be synergistically combined with existing methods to enhance the resilience of LLMs against forgetting.
\end{itemize}

\section{Unveiling the Hidden Nexus: CF and the Model LLS}

We investigate the relationship between CF and the flatness of LLS through three analyses: \textit{i)} \textbf{loss landscape visualization}, \textit{ii)} \textbf{flatness degree of LLS}, and \textit{iii)} \textbf{general task performance},
where the visualized comparison of \textit{i)} and results of \textit{ii) \& iii)} are discussed in \S~\ref{subsec:prob1} and \S~\ref{subsec:prob2}, respectively, to intuitively and quantitatively reveal their correlation.

We simulate continue-learning scenarios with three models: Llama2-7b~\cite{llama} tuned on 1) ``Alpaca~\cite{Alpaca}'', 2) ``Alpaca$\rightarrow$Open-Platypus~\cite{platypus2023}'', and 3) ``Alpaca$\rightarrow$Auto-Wiki~\cite{Auto-wiki:1, Auto-wiki:2}'', where ``$\rightarrow$'' means continue-train on the following dataset and the data gaps are gradually increased. For the experimental details, please refer to the Appendix~\ref{appendix:A.1.1}.

\subsection{Visualized LLS comparison of different CF settings}
\label{subsec:prob1}

Following~\citet{Visualizing,zan-etal-2022-complementarity}, we visualize the 2D loss surface to explore the differences in the LLS of models under different settings. We represent the weight of a well-trained model as $\theta_0$ and define the loss surface with the function: $f(\alpha, \beta)=\mathcal{L}\left(\boldsymbol{\theta}_0+\alpha \boldsymbol{\delta}_1+\beta \boldsymbol{\delta}_2\right),$
where $\mathcal{L}$ is the loss function, $\alpha$ and $\beta$ are scalar values representing the current coordinate, with \bm{${\delta}_1$} and \bm{${\delta}_2$} as two direction vectors. We choose Gaussian noise as the basis vectors for exploring the parameter space, and the axes correspond to $\alpha$ and $\beta$ values.

\begin{table}[t]
\centering
\small
\setlength{\tabcolsep}{1.0mm}
\begin{tabular}{lcccc}
\toprule
&\multicolumn{3}{c}{\textbf{Flatness Degree}} & \textbf{General Perf.}\\
\cline{2-5}
  & {\textit{SC}} & {\textit{AG}} & {\textit{MAG}} & {\textit{MMLU}}\\
\midrule
(a)Alpaca    & 52.87            & 105.37             & 65.53  & 40.53~~~~               \\
(b)$\rightarrow$Open-Platypus & 52.98            & 106.41             & 68.91        &~~~$\text{33.46}^{\textcolor{red}{\downarrow \textbf{-7.1}}}$\\
(c)$\rightarrow$Auto-Wiki     & 53.77            & 111.03             & 70.71  & ~~~~$\text{23.31}^{\textcolor{red}{\downarrow \textbf{-17.2}}}$ \\
\bottomrule
\end{tabular}

\caption{\textbf{Quantitative results of the LLS flatness} (``Flatness Degree'') of LLMs \textbf{and their general task performance} (``General Perf.''). }
\label{table1}

\end{table}

From the LLS visualization in Fig.~\ref{fig:three_images}, we observe that with the continue-learned task gap increases, i.e., (a)$\leq$(b)$\leq$(c), their landscape and contour lines become sharper and more disturbed, \textit{intuitively} demonstrating the \textbf{\textit{high correlation between the flatness degree}} (of model loss landscape) \textbf{\textit{and the forgetting degree}} (of continually tuning LLMs).

\subsection{Quantitative flatness degree and general performance of different CF settings}
\label{subsec:prob2}

Besides intuitive visualization, we quantitatively measure the \textit{flatness degree of loss landscape} (with three metrics: `SC' (Surface Curvature, \%), `AG' (Average Gradient,\%), `MAG' (Mean Absolute Gradient, \%)), and the \textit{model performance on general tasks} (in terms of MMLU benchmark~\cite{MMLU}). Details of our proposed metrics can see Appendix~\ref{appendix:A.3}.

As seen in Tab.~\ref{table1}, for different CF settings (difficulty rank: (a)$\leq$(b)$\leq$(c)), their performance on downstream tasks, i.e., MMLU tasks, is significantly dropped (up to -17.2 on average), confirming our hypothesis that \textbf{\textit{continually training on harder tasks (with larger data gaps) will exacerbate CF}} problem. 
Also, we find that the flatness degree rank is as follows: (a)$\geq$(b)$\geq$(c), where their average values are 74.59, 76.1, and 78.50, respectively, implying that \textbf{\textit{there is a highly positive correlation between the severity of CF and the sharpness of the loss landscape}}.

\section{Improving CF with Sharpness-Aware Optimization}

Motivated by the above findings, we naturally consider introducing the Sharpness-Aware Minimization (SAM,~\citealp{SAM}) method to flatten the optimization landscape, thereby mitigating the CF problem. Given model weights as $w \in \mathbb{R}^d$, loss function as $f$, and a training dataset $S = \{(x_i, y_i)\}_{i=1}^n$ $\boldsymbol{i.i.d.}$ sampled from the distribution $D$, the working mechanism of SAM can be formulated as follows:

\begin{equation}\label{eq1}
\min_{\bm{w}} \max_{\|\bm{\epsilon}\|_2 \leq \rho} f(\bm{w} + \bm{\epsilon}),
\end{equation}

\noindent where \bm{$\epsilon$} is the perturbation in a neighbourhood ball area with a radius denoted as $\rho$. The optimization objective is to ensure that the loss $f$ does not increase substantially with \bm{$\epsilon$} constrained by $\rho$. Further utilizing Taylor series expansion, \bm{$\epsilon$} can be approximate as $\rho \bm{\nabla_w} f(\bm{w})/{\|\bm{\nabla_w} f(\bm{w})\|_2}$. In case, Eq.~\ref{eq1} can be simplified as:

\begin{equation}\label{eq2}
\min_{\bm{w}} f\left(\bm{w} + \rho \frac{\bm{\nabla_w} f(\bm{w})}{\|\bm{\nabla_w} f(\bm{w})\|_2}\right).
\end{equation}

Eq.~\ref{eq2} could be solved through a two-step gradient descent process. Firstly, the perturbation \bm{$\epsilon$} is computed by $\rho \bm{\nabla_w} f(\bm{w})/{\|\bm{\nabla_w} f(\bm{w})\|_2}$. Following this, the second phase of gradient descent performs the actual update of the weights.

\section{Experimental Setting}

\paragraph{Dataset} To validate the capability of SAM on diverse datasets, we conduct experiments on four widely-used instruction fine-tuning datasets: Alpaca~\cite{Alpaca}, ShareGPT52K~\cite{ShareGPT}, MetaMathQA~\cite{metamath}, and Open-Platypus~\cite{platypus2023}. For detailed introduction, please refer to the Appendix~\ref{appendix:A.1.1}.

\paragraph{Model Selection} To further demonstrate the compatibility of SAM with models of different sizes, we conducted experiments on models including TinyLlama-1.1B~\cite{tinyllama}, Llama2-7B, and Llama2-13B~\cite{llama}. 

\begin{table*}[ht]
\centering
\begin{subtable}{\textwidth}
\centering
\small
\setlength{\tabcolsep}{4.5mm}
\begin{tabular}{lcccccc}
\toprule
                                \bf Dataset     &\bf DK  &\bf Understanding &\bf  Reasoning & \bf Exams    &\bf\textit{AVG} & $\bm{\Delta}$   \\ 
\midrule
Alpaca                                     & 40.53 & 58.74         &  63.33 & 45.08    & 51.92  &  --- \\ 
\hdashline
{\color[HTML]{333333} \hspace{1mm} 
 $\rightarrow$ShareGPT(w/o)}      & 26.08 & 52.84         & 58.68 &  45.76    & 45.84 & \textbf{-6.08} \\
{\color[HTML]{333333} \hspace{1mm} 
 $\rightarrow$ShareGPT(w/)}       &  40.08 & 57.91         & 63.78 & 44.41  & 51.55  & \textcolor{red}{\textbf{+5.71}} \\ 
\hdashline
{\color[HTML]{333333} \hspace{1mm}  $\rightarrow$Open-Platypus(w/o)} & 31.13 & 50.70          & 61.09 & 37.97 & 45.22  & \textbf{-6.70}\\
{\color[HTML]{333333} \hspace{1mm}  $\rightarrow$Open-Platypus(w/)}  & 41.07 & 58.28         & 64.50 & 45.08   & 52.23  & \textcolor{red}{\textbf{+7.01}} \\ 
\hdashline
{\color[HTML]{333333} \hspace{1mm}  $\rightarrow$Meta-Math(w/o)}    & 33.13 & 52.61         & 58.46 & 36.27   & 45.12 & \textbf{-6.80}\\
{\color[HTML]{333333} \hspace{1mm}  $\rightarrow$Meta-Math(w/)}      & 34.79 & 55.77         & 62.38 &   42.71  & 48.91 & \textcolor{red}{\textbf{+3.79}} \\ 
\bottomrule
\end{tabular}
\caption{General Performance of Evaluation Tasks on \textbf{SAM (w/o) \& (w/) across Different Datasets}.}
\label{table_res:1}

\end{subtable}

\begin{subtable}{\textwidth}
\centering
\small
\setlength{\tabcolsep}{3.0mm}
\begin{tabular}{llcccccc}
\toprule
\bf Model                      &\bf Dataset                                  &\bf DK  &\bf Understanding   &\bf Reasoning  &\bf Exams  &\bf\textit{AVG} & $\bm{\Delta}$   \\ 
\midrule
\multirow{3}{*}{TinyLlama}  & Alpaca              & 23.16 & 30.62             & 46.16  & 26.78      & 31.68 & ---\\
                            & \hspace{1mm}$\rightarrow$Open-Platypus(w/o) & 23.04 & 31.06    & 46.23    & 27.12               & 31.86  & \textbf{+0.18} \\
                            & \hspace{1mm}$\rightarrow$Open-Platypus(w/)  & 23.14 & 30.39    & 46.91    & 26.10               & 31.64 & \textcolor{red}{\textbf{-0.22}} \\
\hdashline
\multirow{3}{*}{Llama2-7B}  & Alpaca              & 40.53 & 58.74      & 63.33          & 45.08             & 51.92  & --- \\
                            & \hspace{1mm}$\rightarrow$Open-Platypus(w/o) & 31.13 & 50.70   & 61.09     & 37.97              & 45.22  & \textbf{-6.70} \\
                            & \hspace{1mm}$\rightarrow$Open-Platypus(w/)  & 41.07 & 58.28    & 64.50   & 45.08               & 52.23  & \textcolor{red}{\textbf{+7.01}} \\ 
\hdashline
\multirow{3}{*}{Llama2-13B} & Alpaca              & 48.15 & 69.90   & 64.37             & 63.73             & 61.54   & --- \\
                            & \hspace{1mm}$\rightarrow$Open-Platypus(w/o) & 28.23 & 61.92       & 64.12          & 54.58            & 52.21 & \textbf{-9.33}   \\
                            & \hspace{1mm}$\rightarrow$Open-Platypus(w/)  & 49.38 & 69.80        & 65.72        & 63.05             & 61.99 & \textcolor{red}{\textbf{+9.78}}  \\
\bottomrule
\end{tabular}
\caption{General Performance of Evaluation Tasks on \textbf{SAM (w/o) \& (w/) on Different Model Sizes}.}
\label{table_res:2}
\end{subtable}

\begin{subtable}{\textwidth}
\centering
\small
\setlength{\tabcolsep}{4.6mm}
\begin{tabular}{lcccccc}
\toprule
\bf Method            &\bf DK  &\bf Understanding  &\bf Reasoning &\bf Exams  &\bf\textit{AVG} & $\bm{\Delta}$  \\ 
\midrule
Alpaca               & 40.53 & 58.74   & 63.33    & 45.08      & 51.92 & ---\\
\hdashline
\hspace{1mm}$\rightarrow$Open-Platypus(w/o) & 31.13 & 50.70  & 61.09    & 37.97      & 45.22 & \textbf{-6.70}\\
\hspace{1mm}$\rightarrow$Open-Platypus(w/)  & 41.07 & 58.28  & 64.50   & 45.08      & 52.23 & \textcolor{red}{\textbf{+7.01}} \\
\hdashline
Wise-FT (w/o)        & 37.75 & 56.64   & 62.65   & 47.12      & 51.04 & \textbf{-0.88} \\
Wise-FT (w/)         & 40.59 & 58.14  & 64.56  & 44.75      & 52.01  & \textcolor{red}{\textbf{+0.97}} \\
\hdashline
Rehearsal (w/o)       & 33.38 & 54.69  & 61.25   & 43.05      & 48.09 & \textbf{-3.83}  \\
Rehearsal (w/)        & 40.35 & 57.09  & 63.27  & 43.73      & 51.11 & \textcolor{red}{\textbf{+3.02}}\\
\bottomrule
\end{tabular}
\caption{General Performance of Evaluation Tasks on \textbf{SAM Comparing/Combining with Wise-FT and Rehearsal}.}
\label{table_res:3}
\end{subtable}
\caption{\textbf{General Performance of Evaluation Tasks} on Four Analyses of \textbf{SAM}.}
\label{table_res}
\end{table*}

\paragraph{Implementation Details}

For a fair comparison across the same pre-trained language model, our initial phase follows ~\cite{Alpaca} to train an instruction fine-tuning model on the Alpaca dataset. This baseline model underpins our subsequent fine-tuning experiments on various datasets. Besides, all hyper-parameters and settings are maintained uniformly in the following fine-tuning. We opt for the AdamW~\cite{AdamW} optimizer in scenarios without SAM. All experiments are executed on 16 NVIDIA A800 GPUs.

\paragraph{Evaluation Tasks} To effectively assess the extent of the general knowledge retained in LLMs and to quantify the degree of CF, we have implemented a collection of general assessment tasks. These tasks are categorized as follows: (1) \textbf{Domain knowledge benchmark (DK)}: MMLU~\cite{MMLU}, including STEM, Humanities, Social Sciences, and Other domains. (2) \textbf{Reasoning}: SuperGlue~\cite{superglue} include $AX_b$, $AX_g$, RTE, COPA, and datasets like Hellaswag~\cite{hellaswag}, Boolq~\cite{boolq}, and Siqa~\cite{siqa}. (3) \textbf{Understanding}: RACE (middle \& high)~\cite{race}, Openbookqa-fact~\cite{Openbookqa}, and Csl-dev~\cite{csl}. (4) \textbf{Exams}: ARC-c~\cite{arc-c}. Besides, we also conduct experiments on domain-specific datasets like TruthfulQA~\cite{truthfulqa} to assess the degree of retention of specific abilities. Refer to Appendix~\ref{appendix:A.1.2} for more details.

\section{Experimental Results}

We demonstrate the superiority of our proposed method through four experimental analyses:
\begin{enumerate}
    \item Mitigating CF on different datasets
    \item Mitigating CF across different model sizes
    \item Comparison with other advanced methods
    \item Complementary to existing works
\end{enumerate}

The average results for 4 evaluation categories are presented in Tab.~\ref{table_res} (values in \%), where ``(w/o)'' and ``(w/)'' refers to train without and with SAM, respectively. The four experiments are as follows, and the total experimental results can be refereed to the Appendix~\ref{appendix:A.6}.

\paragraph{Performance of SAM in Mitigating CF on different datasets} As presented in Tab.~\ref{table_res}(a), for different CF settings (Alpaca as the baseline, difficulty rank: ShareGPT52K$\leq$Open-Platypus$\leq$MetaMathQA), their performance on evaluation tasks experiences a significant decline ($\Delta$ CF: ShareGPT: 11.71\%; Open-Platypus: 12.90\%; Meta-Math: 13.10\%), confirming that different degrees of CF occur in the fine-tuning phase. \textbf{\textit{With the introduction of SAM, the CF is markedly mitigated}} and the performance is comparable to baseline. Notably, the Tab.~\ref{table_res} shows the results in balanced training cost, i.e., (w/o) epoch=2 and (w/) epoch=1. For complete epoch experiments and experiment of capability preseving in domain, refer to the Appendix~\ref{appendix:A.4} and Appendix~\ref{appendix:finetuning_domain}, respectively.

\paragraph{Performance of SAM on LLMs of Different Sizes} For the results in Tab.~\ref{table_res}(b), we conduct Alpaca$\rightarrow$Open-Platypus across models of different sizes, and they exhibited different degrees of performance dropping (e.g., $\Delta$ DK: TinyLlama-1.1B: 0.52\%; Llama2-7B: 23.19\%; Llama2-13B: 41.37\%), indicating that \textbf{\textit{the severity of the forgetting increasing with model size}}. Compared to the performance of SAM in mitigating CF across models, \textbf{\textit{SAM could significantly mitigate the extent of CF with model size increasing}}.

\paragraph{In Comparison with Other Advanced Anti-Forgetting Methods} For the comparative experimental results represents in Tab.\ref{table_res}(c), \textbf{\textit{SAM significantly outperforms Wise-FT and Rehearsal}}, although the other two also achieve decent performance ($\Delta$ Mitigating CF: SAM: 7.01\%; Wise-FT: 5.82\%; Rehearsal: 3.79\%).

\paragraph{Complementary to Existing Advanced Anti-Forgetting Methods} For the results in Table \ref{table_res}(c), SAM could effectively improve the performance of CF along with the two methods ($\Delta$ Mitigating CF: Wise-FT(w/): 0.97\%; Rehearsal(w/): 3.02\%), which illustrates that \textbf{\textit{SAM can be orthogonally combined with other methods}}, providing incremental benefits to mitigate CF.

\section{Conclusion}
We reexamine the phenomenon of Catastrophic Forgetting (CF) and reveal the direct link between the flatness of the model loss landscape (LLS) and the extent of CF in LLMs. Based on our findings, we introduce sharpness-aware optimization to flatten the LLS, thereby mitigating CF. Experiments show its effectiveness, especially in larger models, suggesting our method may become a standard strategy to mitigate CF during LLMs tuning.

In future work, we would like to explore more strategies to orthogonally address the forgetting problem in LLM tuning, such as advanced optimizers~\cite{sun2024adasam,zhong-etal-2022-improving,yang2023data}, proper anti-forgetting learning curricula~\cite{ding2021progressive,zhou2021self,faber2024mnist}, and model merging techniques~\cite{li2023deep,he2023merging}.

\section*{Limitations}
Despite the progress we made, there still exist limitations in our work. On the one hand, our experiments are concentrated on a specific aspect of the Catastrophic Forgetting (CF) --- the direct link between the flatness of the model's loss landscape and CF. While this finding makes us introduce the sharpness-aware minimization technique to flatten the loss landscape and thereby mitigate CF, it also means that other potentially influential factors in CF have not been comprehensively explored. On the other hand, our research primarily addresses CF during the fine-tuning phase of LLMs, but it does not directly tackle CF that may occur during other stages. For instance, CF can also be a significant challenge when models are updated with new data post-deployment. Therefore, the scope of our findings and the applicability of our solution may be limited to specific phases of the LLMs lifecycle.

\section*{Ethics Statement}
We take ethical considerations seriously and strictly adhere to the ACL Ethics Policy. This paper revisits a major obstacle of fine-tuning the large language models -- Catastrophic Forgetting, and provides an optimization-level solution to alleviate it. All employed models and datasets in this paper are publicly available and have been widely adopted by researchers. All experimental results upon these open models and datasets are reported accurately and objectively. Thus, we believe that this research will not pose any ethical issues.

\paragraph{Reproducibility.} We provide the detailed experimental setup in the Appendix, such as hyper-parameters and statistic descriptions. Also, we release code to facilitate reproducing the experimental results: \url{https://github.com/Li-Hyn/LLM_CatastrophicForgetting}.

\bibliography{arxiv_version}

\appendix

\section{Appendix}
\label{sec:appendix}

\subsection{Details of Training and Evaluation Datasets}
\label{appendix:A.1}

As outlined in \S 4 and \S 2, our study involves extensive experiments across five widely-utilized datasets. Additionally, we assess our model's performance on selected tasks from the SuperGLUE benchmark along with other distinct datasets. In this section, we provide detailed descriptions of the datasets and tasks employed. The description of each task includes:

\subsubsection{Training Datasets}
\label{appendix:A.1.1}
\begin{enumerate}
    \item \textbf{Alpaca}: Alpaca~\cite{Alpaca} is a dataset of 52,000 instructions and demonstrations generated by OpenAI's text-davinci-003 engine. This instruction data can be used to conduct instruction-tuning for language models and make the language model follow instruction better.
    \item \textbf{ShareGPT52K}: ShareGPT52K~\cite{ShareGPT} is a collection of approximately 52,000 conversations scraped via the ShareGPT API. These conversations include both user prompts and responses from OpenAI's ChatGPT.
    \item \textbf{MetaMathQA}: MetaMathQA~\cite{metamath} is a dataset of 395,000 instructions which augmented from the training sets of GSM8K~\cite{cobbe2021gsm8k} and MATH~\cite{math}.
    \item \textbf{Open-Platypus}: Open-Platypus~\cite{platypus2023} is focused on improving LLM logical reasoning skills and has the item amount to 24,900.
    \item \textbf{Auto-Wiki}: Auto-Wiki~\cite{Auto-wiki:1, Auto-wiki:2} provides a set of aligned sentences from English Wikipedia and Simple English Wikipedia as a resource to train sentence simplification systems.

\end{enumerate}

\begin{table*}[ht]
\centering
\small
\setlength{\tabcolsep}{4.3mm}
\begin{tabular}{clccccc}
\toprule
\bf{Epoch}                             & \bf{Dataset}                                             & \bf{DK}       & \bf{Understanding}            & \bf{Reasoning}                                                      & \bf{Exams}                                  & \bf\textit{AVG}                                  \\
\midrule
3                                 & Alpaca                                              & 40.53      & 58.74                  & 63.33                                                                  & 45.08                                 & 51.92                                 \\
\hdashline
2                                 & {\color[HTML]{000000} \hspace{1mm}$\rightarrow$ShareGPT(w/o)}               & 26.08               & 52.84           & 58.68                                                                 & 45.76                                 & 45.84                                 \\
{\color[HTML]{3166FF} \textbf{1}} & {\color[HTML]{3166FF} \hspace{1mm}\textbf{$\rightarrow$ShareGPT(w/o)}}      & {\color[HTML]{3166FF} \textbf{24.98}} & {\color[HTML]{3166FF} \textbf{52.40}}  & {\color[HTML]{3166FF} \textbf{59.12}}  & {\color[HTML]{3166FF} \textbf{45.08}} & {\color[HTML]{3166FF} \textbf{45.39}} \\
1                                 & {\color[HTML]{000000} \hspace{1mm}$\rightarrow$ShareGPT(w/)}                & 40.08          & 57.91                   & 63.78                                                               & 44.41                                 & {\color[HTML]{FE0000} \textbf{51.55}} \\
\hdashline
2                                 & {\color[HTML]{000000} \hspace{1mm}$\rightarrow$Open-Platypus(w/o)}          & 31.13            & 50.70                & 61.09                                                              & 37.97                                 & 45.22                                 \\
{\color[HTML]{3166FF} \textbf{1}} & {\color[HTML]{3166FF} \textbf{\hspace{1mm}$\rightarrow$Open-Platypus(w/o)}} & {\color[HTML]{3166FF} \textbf{29.89}} & {\color[HTML]{3166FF} \textbf{49.37}}  & {\color[HTML]{3166FF} \textbf{60.60}}  & {\color[HTML]{3166FF} \textbf{40.34}} & {\color[HTML]{3166FF} \textbf{45.05}} \\
1                                 & {\color[HTML]{000000} \hspace{1mm}$\rightarrow$Open-Platypus(w/)}           & 41.07                 & 58.28         & 64.50                                                                & 45.08                                 & {\color[HTML]{FE0000} \textbf{52.23}} \\
\hdashline
2                                 & {\color[HTML]{000000} \hspace{1mm}$\rightarrow$Meta-Math(w/o)}              & 33.13         & 52.61                 & 58.46                                                                 & 36.27                                 & 45.12                                 \\
{\color[HTML]{3166FF} \textbf{1}} & {\color[HTML]{3166FF} \textbf{\hspace{1mm}$\rightarrow$Meta-Math(w/o)}}     & {\color[HTML]{3166FF} \textbf{32.84}} & {\color[HTML]{3166FF} \textbf{52.69}}    & {\color[HTML]{3166FF} \textbf{58.43}}  & {\color[HTML]{3166FF} \textbf{38.64}} & {\color[HTML]{3166FF} \textbf{45.65}} \\
1                                 & {\color[HTML]{000000} \hspace{1mm}$\rightarrow$Meta-Math(w/)}               & 34.79            & 55.77             & 62.38                                                                 & 42.71                                 & {\color[HTML]{FE0000} \textbf{48.91}} \\
\bottomrule
\end{tabular}
\caption{General Performance of \textbf{Different Epochs on Llama2-7b}.}
\label{table_epoch}
\end{table*}

\begin{table}[ht]
\small
\setlength{\tabcolsep}{0.35mm}
\centering
\begin{tabular}{lccc}
\toprule
\bf Dataset             &\bf Hellaswag &\bf TruthfulQA &\bf\textit{AVG}~~~~   \\ 
\midrule
Alpaca              & 65.00     & 22.64       & 43.82~~~~ \\
\hspace{1mm}$\rightarrow$Open-Platypus(w/o) & 73.16     & 23.87       & ~~~$\text{48.52}^{\textcolor{red}{\uparrow \textbf{+4.7}}}$ \\
\hspace{1mm}$\rightarrow$Open-Platypus(w/)  & 73.49     & 23.99       & ~~~$\text{48.74}^{\textcolor{red}{\uparrow \textbf{+4.9}}}$ \\ 
\bottomrule
\end{tabular}
\caption{\textbf{Performance in Fine-tuning Domain}.}
\label{table3}
\end{table}

\subsubsection{Evaluation Tasks}
\label{appendix:A.1.2}
\begin{enumerate}
    \item \textbf{MMLU}: Massive Multitask Language Understanding benchmark (MMLU)~\cite{MMLU} to evaluate the knowledge stored in the LLMs, which can be divided into STEM, Human, Social, and Other.
    \item \bm{$AX_b$}: AX-b~\cite{superglue} is a broad-coverage diagnostic task, which requires to determine the logical relation between the given sentence pair, with three relations: entailment, contradiction and neutral. 
    \item \bm{$AX_g$}: AX-g~\cite{superglue} is a Winogender diagnostic task, which requires to determine which noun the pronoun refers to according to the given sentence and pronoun.
    \item \textbf{RTE}: Recognizing Textual Entailment~\cite{RTE}, given a premise and a hypothesis, is a task to predict whether the premise entails the hypothesis. 
    \item \textbf{COPA}: COPA~\cite{superglue} is a causal inference task, which requires to select the correct causal relation based on the given premise.
    \item \textbf{Hellaswag}: HellaSwag~\cite{hellaswag} is a challenge dataset for evaluating commonsense natural language inference, which is specially hard for state-of-the-art models, though its questions are trivial for humans (>95\% accuracy).
    \item \textbf{Boolq}: Boolean Question~\cite{boolq} is a question-answering task where each sample consists of a short passage and a yes/no question about the passage.
    \item \textbf{Siqa}: Siqa~\cite{siqa} is a social interaction question-answering task, which requires selecting the most reasonable behaviour based on the given scenario and three possible subsequent behaviours.
    \item \textbf{RACE}: RACE~\cite{race} is a large-scale reading comprehension dataset with more than 28,000 passages and nearly 100,000 questions. 
    \item \textbf{Openbookqa-fact}: OpenBookQA~\cite{Openbookqa} contains questions that require multi-step reasoning, application of common-sense knowledge, and in-depth comprehension of text.
    \item \textbf{Csl-dev}: CSL~\cite{csl} is a large-scale Chinese Scientific Literature dataset, which contains the titles, abstracts, keywords and academic fields of 396k papers
    \item \textbf{ARC-c}: The AI2’s Reasoning Challenge (ARC)~\cite{arc-c} is a multiple-choice question-answering dataset, containing questions from science exams from grade 3 to grade 9.
    \item \textbf{TruthfulQA}: TruthfulQA~\cite{truthfulqa} is a benchmark to measure whether a language model is truthful in generating answers to questions.

\end{enumerate}

\subsection{Experimental Setup}
\label{appendix:A.2}
In this paper, we trained from the base model (TinyLlama, Llama2-7b, and Llama2-13B). For a fair comparison, we follow the setup in~\cite{Alpaca} in all first phases of experiments, that is, the number of epochs is 3, and the learning rate (lr) is set as 2e-5, with the batch size of 128. We confirmed that the Alpaca model, trained under this configuration, exhibits performance roughly equivalent to that of the original model as reported in the~\cite{Alpaca}.

For the subsequent training procedures, we aligned our fine-tuning configuration with that used in real-world downstream tasks. Taking into account the characteristic of SAM performing two forward passes, to facilitate a fair comparison, we established our experimental parameters as follows: without SAM: epochs = 2, lr = 5e-6, batch size = 128, and the optimizer is AdamW; with SAM: epochs = 1, lr = 5e-6, batch size = 128. For the selection of the $\rho$ parameter within the SAM training process, we suggest a reference value of $\rho = 2$.


\begin{table*}[ht]
\centering
\begin{subtable}{\textwidth}
\centering
\small
\setlength{\tabcolsep}{5mm}
\begin{tabular}{lcccccc}
\toprule
\bf{Epoch}                    & \bf{Dataset}                                                                        & \bf{STEM}                         & \bf{Humanities}                   & \bf{Social Sciences}              & \bf{Other}                        \\
\midrule
3                        & Alpaca                                                             & 31.60                        & 43.03                        & 47.23                        & 43.94                        \\
\hdashline
2                        & \hspace{1mm} $\rightarrow$ShareGPT(w/o)                                            & 22.69                        & 29.72                        & 23.57                        & 29.20                        \\
{\color[HTML]{3166FF} 1} & {\color[HTML]{3166FF} \hspace{1mm} $\rightarrow$ShareGPT(w/o)}    & {\color[HTML]{3166FF} 22.55} & {\color[HTML]{3166FF} 30.35} & {\color[HTML]{3166FF} 22.96} & {\color[HTML]{3166FF} 24.83} \\
1                        & \hspace{1mm} $\rightarrow$ShareGPT(w/)                                              & 31.08                        & 42.82                        & 47.24                        &  42.96 \\
\hdashline
2                        & \hspace{1mm} $\rightarrow$Open-Platypus(w/o)                                             & 25.44                        & 34.70                        & 34.20                        & 32.50                        \\
{\color[HTML]{3166FF} 1} & {\color[HTML]{3166FF} \hspace{1mm} $\rightarrow$Open-Platypus(w/o)} & {\color[HTML]{3166FF} 24.95} & {\color[HTML]{3166FF} 33.77} & {\color[HTML]{3166FF} 31.58} & {\color[HTML]{3166FF} 31.21} \\
1                        & \hspace{1mm} $\rightarrow$Open-Platypus(w/)                                           & 32.18                        & 43.56                        & 48.11                        &  44.15 \\
\hdashline
2                        & \hspace{1mm} $\rightarrow$Meta-Math(w/o)                                               & 26.57                        & 34.77                        & 39.93                        & 34.19                        \\
{\color[HTML]{3166FF} 1} & {\color[HTML]{3166FF} \hspace{1mm} $\rightarrow$Meta-Math(w/o)}   & {\color[HTML]{3166FF} 28.22} & {\color[HTML]{3166FF} 33.45} & {\color[HTML]{3166FF} 37.74} & {\color[HTML]{3166FF} 34.03} \\
1                        & \hspace{1mm} $\rightarrow$Meta-Math(w/)                                              & 28.48                        & 34.84                        & 39.50                        &  38.82 \\
\bottomrule
\end{tabular}
\caption{\textbf{DK Performance} of Different Epochs on Llama2-7b.}
\label{table_epoch_multitask:DK}
\end{subtable}

\begin{subtable}{\textwidth}
\centering
\small
\setlength{\tabcolsep}{2.5mm}
\begin{tabular}{clccccc}
\toprule
                         &                                             & \multicolumn{4}{c}{\bf{Understanding}}                                                                                        & \bf{Exams}                        \\
                         \cline{3-7}
\multirow{-2}{*}{\bf{Epoch}}  & \multirow{-2}{*}{\bf{Dataset}}                   & RACE (middle)                & RACE (high)                  & Openbookqa-fact             & Csl-dev                      & ARC-c                        \\
\midrule
3                        & Alpaca                              & 55.22                        & 55.22                        & 66.4                        & 58.13                        & 45.08                        \\
\hdashline
2                        & \hspace{1mm} $\rightarrow$ShareGPT(w/o)                             & 50.7                         & 47.4                         & 57                          & 56.25                        & 45.76                        \\
{\color[HTML]{3166FF} 1} & {\color[HTML]{3166FF} \hspace{1mm} $\rightarrow$ShareGPT(w/o)}      & {\color[HTML]{3166FF} 49.86} & {\color[HTML]{3166FF} 47.74} & {\color[HTML]{3166FF} 57}   & {\color[HTML]{3166FF} 55}    & {\color[HTML]{3166FF} 45.08} \\
1                        & \hspace{1mm} $\rightarrow$ShareGPT(w/)               & 55.43                        & 51.49                        & 66.6                        & 58.13                        & 44.41                        \\
\hdashline
2                        & \hspace{1mm} $\rightarrow$Open-Platypus(w/o)                        & 46.31                        & 44.54                        & 53.8                        & 58.13                        & 37.97                        \\
{\color[HTML]{3166FF} 1} & {\color[HTML]{3166FF} \hspace{1mm} $\rightarrow$Open-Platypus(w/o)} & {\color[HTML]{3166FF} 46.52} & {\color[HTML]{3166FF} 44.17} & {\color[HTML]{3166FF} 51.8} & {\color[HTML]{3166FF} 55}    & {\color[HTML]{3166FF} 40.34} \\
1                        & \hspace{1mm} $\rightarrow$Open-Platypus(w/)                         & 56.34                        & 51.69                        & 67.6                        & 57.5                         & 45.08                        \\
\hdashline
2                        & \hspace{1mm} $\rightarrow$Meta-Math(w/o)                            & 56.55                        & 51.8                         & 45.2                        & 56.88                        & 36.27                        \\
{\color[HTML]{3166FF} 1} & {\color[HTML]{3166FF} \hspace{1mm} $\rightarrow$Meta-Math(w/o)}     & {\color[HTML]{3166FF} 54.67} & {\color[HTML]{3166FF} 52.03} & {\color[HTML]{3166FF} 47.8} & {\color[HTML]{3166FF} 56.25} & {\color[HTML]{3166FF} 38.64} \\
1                        & \hspace{1mm} $\rightarrow$Meta-Math(w/)                             & 53.2                         & 49.69                        & 65.2                        & 55                           & 42.71    \\
\bottomrule
\end{tabular}
\caption{\textbf{Performance of Understanding and Exams} Different Epochs on Llama2-7b.}
\label{table_epoch_multitask:Under}
\end{subtable}

\begin{subtable}{\textwidth}
\centering
\small
\setlength{\tabcolsep}{2.8mm}
\begin{tabular}{clccccccc}
\toprule
    \bf{Epoch}                    & \bf{Dataset}                                                                             & \bf{BoolQ}                        & \bf{AX\_b}                        & \bf{AX\_g}                        & \bf{ RTE}   & \bf{COPA}  & \bf{Hellaswag}                    & \bf{Siqa}                         \\
\midrule
3                        & Alpaca                                                                      & 78.29                        & 59.42                        & 60.67                        & 61.37                        & 65.00                        & 65.00                        & 53.53                        \\
\hdashline
2                        & \hspace{1mm} $\rightarrow$ShareGPT(w/o)                             & 75.20                        & 57.70                        & 50.00                        & 47.65                        & 60.00                        & 73.81                        & 46.42                        \\
{\color[HTML]{3166FF} 1} & {\color[HTML]{3166FF} \hspace{1mm} $\rightarrow$ShareGPT(w/o)}      & {\color[HTML]{3166FF} 73.52} & {\color[HTML]{3166FF} 57.88} & {\color[HTML]{3166FF} 50.00} & {\color[HTML]{3166FF} 49.46} & {\color[HTML]{3166FF} 64.00} & {\color[HTML]{3166FF} 72.48} & {\color[HTML]{3166FF} 46.52} \\
1                        & \hspace{1mm} $\rightarrow$ShareGPT(w/)                              & 77.92                        & 59.33                        & 56.46                        & 61.37                        & 65.00                        & 73.18                        & 53.22                        \\
\hdashline
2                        & \hspace{1mm} $\rightarrow$Open-Platypus(w/o)                        & 78.20                        & 59.24                        & 50.00                        & 50.18                        & 69.00                        & 73.16                        & 47.85                        \\
{\color[HTML]{3166FF} 1} & {\color[HTML]{3166FF} \hspace{1mm} $\rightarrow$Open-Platypus(w/o)} & {\color[HTML]{3166FF} 75.72} & {\color[HTML]{3166FF} 58.70} & {\color[HTML]{3166FF} 50.00} & {\color[HTML]{3166FF} 49.10} & {\color[HTML]{3166FF} 70.00} & {\color[HTML]{3166FF} 72.29} & {\color[HTML]{3166FF} 48.36} \\
1                        & \hspace{1mm} $\rightarrow$Open-Platypus(w/)                         & 79.11                        & 59.06                        & 63.76                        & 58.48                        & 65.00                        & 73.49                        & 52.61                        \\
\hdashline
2                        & \hspace{1mm} $\rightarrow$Meta-Math(w/o)                            & 69.60                        & 58.42                        & 50.00                        & 47.65                        & 64.00                        & 72.09                        & 47.49                        \\
{\color[HTML]{3166FF} 1} & {\color[HTML]{3166FF} \hspace{1mm} $\rightarrow$Meta-Math(w/o)}     & {\color[HTML]{3166FF} 70.61} & {\color[HTML]{3166FF} 58.33} & {\color[HTML]{3166FF} 50.00} & {\color[HTML]{3166FF} 47.65} & {\color[HTML]{3166FF} 62.00} & {\color[HTML]{3166FF} 71.90} & {\color[HTML]{3166FF} 48.52} \\
1                        & \hspace{1mm} $\rightarrow$Meta-Math(w/)                             & 77.61                        & 58.88                        & 53.65                        & 53.43                        & 70.00                        & 71.70                        & 51.43   \\
\bottomrule
\end{tabular}
\caption{\textbf{Reasoning Performance} of Different Epochs on Llama2-7b.}
\label{table_epoch_multitask:reasoning}
\end{subtable}

\caption{ Comprehensive Experimental Results of \textbf{Different Epochs on Llama2-7b}.}
\label{table_epoch_multitask}

\end{table*}

\subsection{Flatness Metric}
\label{appendix:A.3}
Consider a function $f(x_i,y_j)$, where $x_i$ and $y_j$ represent two arbitrary parameters $param_1$ and $param_2$, and the N represents the total number of points in this area.

\noindent \textbf{Surface Curvature}\\
Surface curvature (SC) typically refers to a measure that describes the degree of curvature of a surface at a certain point or area. A lower average curvature usually indicates a flatter loss landscape.
To quantify the average curvature of a loss surface defined by two parameters, the approximate quantification of the curvature of the function $f(x_i,y_j)$ within its parameter space is as follows:

\begin{equation}
K\left(x_i, y_j\right)=f_{x x}\left(x_i, y_j\right)+f_{y y}\left(x_i, y_j\right).
\end{equation}

We define the overall curvature as the average curvature, which is computed by taking the mean of the absolute values of the curvature across the entire parameter space, formulated as:

\begin{equation}
SC(x, y)=\frac{1}{N^2} \sum_{i=1}^N \sum_{j=1}^N\left|K\left(x_i, y_j\right)\right|.
\end{equation}

\noindent \textbf{Average Gradient}\\
We take average gradient (AG) serve as an indicator of the flatness of loss landscape, with smaller AG typically corresponding to flatter regions of the loss landscape. The AG refers to the average of the gradient vectors calculated within a certain region of the function, expressed as follows:

\begin{equation}
\text { AG }=\frac{1}{N} \sum_{i=1}^N \nabla f\left(x_i, y_i\right).
\end{equation}

\noindent \textbf{Mean Absolute Gradient}\\
To more comprehensively evaluate the flatness of LLS, we introduce the metric Mean Absolute Gradient (MAG). With lower MAG, the loss landscape tends to be flatter. MAG quantifies the average rate of change of the loss function in its parameter space. The function could be expressed as follows:

\begin{equation}
\text { MAG }=\frac{1}{N-1} \sum_{i=2}^N\left|t_i-t_{i-1}\right|,
\end{equation}

\noindent where $t_i$ represents any data point in LLS.




\subsection{Experiment on different Epochs}
\label{appendix:A.4}

Given the need for a fair comparison, the number of epochs for training differs between SAM without (w/o) and with (w/) settings. Consequently, it is imperative to fully present the results for SAM (w/o) at epoch = 1, to substantiate the rigor of our experimental methodology.

The results in Tab.~\ref{table_epoch} clearly show that in most cases the results at epoch=2 are superior to those at epoch=1. Furthermore, across all three settings of the same dataset, \textbf{\textit{SAM (w/) consistently achieved the best results}}. These experimental results further demonstrate the superiority of our method. Also, to further present the result, we give the complete results of all datasets in Tab.~\ref{table_epoch_multitask}.

\begin{table*}[ht]
\centering

\begin{subtable}{\textwidth}
\centering
\small
\setlength{\tabcolsep}{4.3mm}
\begin{tabular}{llcccc}
\toprule
\bf{Models}         & \bf{Dataset}                         & \bf{STEM}  & \bf{Humanities} & \bf{Social Sciences} & \bf{Other} \\
\midrule
\multirow{3}{*}{TinyLlama}  & Alpaca                          & 22.18 & 24.03      & 22.88           & 23.84 \\
               & \hspace{1mm} $\rightarrow$Open-platypus(w/o) & 22.29 & 23.55      & 22.59           & 23.90 \\
               & \hspace{1mm} $\rightarrow$Open-platypus(w/)  & 21.96 & 24.16      & 22.53           & 23.92 \\
\hdashline
\multirow{3}{*}{Llama2-13b}     & Alpaca                          & 39.58 & 52.29      & 55.83           & 48.75 \\
               & \hspace{1mm} $\rightarrow$Open-platypus(w/o) & 24.58 & 33.23      & 27.42           & 28.98 \\
               & \hspace{1mm} $\rightarrow$Open-platypus(w/)  & 39.51 & 55.59      & 56.92           & 49.84    \\
\bottomrule
\end{tabular}
\caption{\textbf{Dk Performance} of Different sizes of LLMs.}
\label{table_model_size:DK}
\end{subtable}

\begin{subtable}{\textwidth}
\centering
\small
\setlength{\tabcolsep}{1.9mm}
\begin{tabular}{llccccc}
\toprule
\multirow{2}{*}{\bf{Models}}     & \multirow{2}{*}{\bf{Dataset}} & \multicolumn{4}{c}{\bf{Understanding}}                       & \bf{Exams} \\
\cline{3-7}
                            &                          & RACE (middle) & RACE (high) & Openbookqa-fact & Csl-dev & ARC-c \\
                            \midrule
\multirow{3}{*}{TinyLlama}  & Alpaca                   & 22.91         & 21.90       & 25.80           & 51.88   & 26.78 \\
                            & \hspace{1mm} $\rightarrow$Open-Platypus(w/o)     & 24.09         & 22.27       & 26.00           & 51.88   & 27.12 \\
                            & \hspace{1mm} $\rightarrow$Open-Platypus(w/)       & 22.63         & 21.84       & 25.2            & 51.88   & 26.10 \\
                            \hdashline
\multirow{3}{*}{Llama2-13b} & Alpaca                   & 72.84         & 68.21       & 80.40           & 58.13   & 63.73 \\
                            & \hspace{1mm} $\rightarrow$Open-Platypus(w/o)     & 71.10         & 68.67       & 50.40           & 57.5    & 54.58 \\
                            & \hspace{1mm} $\rightarrow$Open-Platypus(w/)      & 72.49         & 68.15       & 79.8            & 58.75   & 63.05  \\
\bottomrule
\end{tabular}
\caption{\textbf{Performance of Understanding and Exams} of Different sizes of LLMs.}
\label{table_model_size:under}
\end{subtable}

\begin{subtable}{\textwidth}
\centering
\small
\setlength{\tabcolsep}{2.5mm}
\begin{tabular}{llccccccc}
\toprule
\bf{Models}                      & \bf{Dataset}              & \bf{BoolQ} & \bf{AX\_b} & \bf{AX\_g} & \bf{RTE}   & \bf{COPA}  & \bf{Hellaswag} & \bf{Siqa}  \\
\midrule
\multirow{3}{*}{TinyLlama}  & Alpaca               & 43.98 & 45.56 & 50.28 & 52.71 & 53.00 & 43.91     & 33.67 \\
                            & \hspace{1mm} $\rightarrow$ Open-Platypus(w/o) & 45.47 & 49.28 & 50.00 & 50.90 & 53.00 & 42.86     & 32.09 \\
                            & \hspace{1mm} $\rightarrow$Open-Platypus(w/)   & 44.50 & 52.99 & 50.28 & 53.79 & 52.00 & 42.75     & 32.09 \\
                            \hdashline
\multirow{3}{*}{Llama2-13b} & Alpaca               & 85.41 & 52.54 & 55.34 & 49.46 & 68.00 & 77.61     & 62.23 \\
                            & \hspace{1mm} $\rightarrow$Open-Platypus(w/o) & 83.27 & 56.07 & 55.90 & 48.38 & 72.00 & 77.17     & 56.04 \\
                            & \hspace{1mm} $\rightarrow$Open-Platypus(w/)  & 85.57 & 52.63 & 55.62 & 50.18 & 68.00 & 85.53     & 62.54  \\
\bottomrule
\end{tabular}
\caption{\textbf{Reasoning Performance} of Different sizes of LLMs.}
\label{table_model_size:reasoning}
\end{subtable}

\caption{ Comprehensive Experimental Results across \textbf{Different sizes of LLMs}.}
\label{table_size_multitask}
\end{table*}

\begin{table*}[ht]
\centering

\begin{subtable}{\textwidth}
\centering
\small
\setlength{\tabcolsep}{4.0mm}
\begin{tabular}{llcccc}
\toprule
\bf{Method}         & \bf{Dataset}                               & \bf{STEM}  & \bf{Humanities}  & \bf{Social Sciences} & \bf{Other} \\
\midrule
Wife-FT(w/o)   & \multirow{2}{*}{\hspace{1mm} Alpaca$\rightarrow$Open-Platypus} & 29.95 & 41.04      & 42.77           & 40.42 \\
Rehearsal(w/o) &                                       & 26.98 & 35.00      & 38.32           & 35.90  \\
\bottomrule
\end{tabular}
\caption{\textbf{DK Performance} of Comparison Methods.}
\label{table_compare_multitask:DK}
\end{subtable}

\begin{subtable}{\textwidth}
\centering
\small
\setlength{\tabcolsep}{1.8mm}
\begin{tabular}{llccccc}
\toprule
\multirow{2}{*}{\bf{Models}}     & \multirow{2}{*}{\bf{Dataset}} & \multicolumn{4}{c}{\bf{Understanding}}                       & \bf{Exams} \\
\cline{3-7}
                            &                          & RACE (middle) & RACE (high) & Openbookqa-fact & Csl-dev & ARC-c \\
                            \midrule
Wife-FT(w/o)   & \multirow{2}{*}{Alpaca$\rightarrow$Open-Platypus} & 53.48         & 50.97       & 64.60           & 57.50   & 47.12 \\
Rehearsal(w/o) &                                & 53.13         & 50.74       & 58.00           & 56.88   & 43.05  \\
\bottomrule
\end{tabular}
\caption{\textbf{Performance of Understanding and Exams} of Comparison Methods.}
\label{table_compare_multitask:under}
\end{subtable}

\begin{subtable}{\textwidth}
\centering
\small
\setlength{\tabcolsep}{2.3mm}
\begin{tabular}{llccccccc}
\toprule
\bf{Method}         & \bf{Dataset}                        &  \bf{BoolQ} & \bf{AX\_b} & \bf{AX\_g} & \bf{RTE}   & \bf{COPA}  & \bf{Hellaswag} & \bf{Siqa}  \\
\midrule
Wife-FT(w/o)   & \multirow{2}{*}{Alpaca$\rightarrow$Open-Platypus} & 77.77 & 59.24 & 51.97 & 57.76 & 64.00 & 73.96     & 53.84 \\
Rehearsal(w/o) &                                & 78.01 & 60.78 & 50.00 & 51.99 & 66.00 & 73.43     & 48.52  \\
\bottomrule
\end{tabular}
\caption{\textbf{Reasoning Performance} of Comparison Methods.}
\label{table_compare_multitask:reasoning}
\end{subtable}
\caption{ Comprehensive Experimental Results of \textbf{Comparison with Other Methods}.}
\label{table_compare_multitask}
\end{table*}

\begin{table*}[ht]
\centering

\begin{subtable}{\textwidth}
\centering
\small
\setlength{\tabcolsep}{4.0mm}
\begin{tabular}{llcccc}
\toprule
\bf{Method}         & \bf{Dataset}                               & \bf{STEM}  & \bf{Humanities}  & \bf{Social Sciences} & \bf{Other} \\
\midrule
Wife-FT(w/)   & \multirow{2}{*}{\hspace{1mm} Alpaca$\rightarrow$Open-Platypus} & 31.81 & 43.1      & 47.3           & 43.8 \\
Rehearsal(w/) &                                       & 32.13 & 42.53      & 46.97           & 43.19  \\
\bottomrule
\end{tabular}
\caption{\textbf{DK Performance} of Methods in Combination with SAM.}
\label{table_combine_multitask:DK}
\end{subtable}

\begin{subtable}{\textwidth}
\centering
\small
\setlength{\tabcolsep}{1.8mm}
\begin{tabular}{llccccc}
\toprule
\multirow{2}{*}{\bf{Models}}     & \multirow{2}{*}{\bf{Dataset}} & \multicolumn{4}{c}{\bf{Understanding}}                       & \bf{Exams} \\
\cline{3-7}
                            &                          & RACE (middle) & RACE (high) & Openbookqa-fact & Csl-dev & ARC-c \\
                            \midrule
Wife-FT(w/)   & \multirow{2}{*}{Alpaca$\rightarrow$Open-Platypus} & 56.06         & 51.6       & 67.40           & 57.50   & 44.75 \\
Rehearsal(w/) &                                & 54.25          & 50.40       & 66.20            & 57.50    & 43.73  \\
\bottomrule
\end{tabular}
\caption{\textbf{Performance of Understanding and Exams} in Combination with SAM.}
\label{table_combine_multitask:under}

\end{subtable}

\begin{subtable}{\textwidth}
\centering
\small
\setlength{\tabcolsep}{2.3mm}
\begin{tabular}{llccccccc}
\toprule
\bf{Method}         & \bf{Dataset}                        &  \bf{BoolQ} & \bf{AX\_b} & \bf{AX\_g} & \bf{RTE}   & \bf{COPA}  & \bf{Hellaswag} & \bf{Siqa}  \\
\midrule
Wife-FT(w/)   & \multirow{2}{*}{Alpaca$\rightarrow$Open-Platypus} & 78.69 & 58.97 & 60.96 & 61.73 & 65.00 & 73.67     & 52.87 \\
Rehearsal(w/) &                                & 76.70  & 57.97  & 55.62  & 60.65  & 67.00 & 72.75     & 52.20   \\
\bottomrule
\end{tabular}
\caption{\textbf{Reasoning Performance} in Combination with SAM.}
\label{table_combine_multitask:reasoning}
\end{subtable}

\caption{ Comprehensive Experimental Results of \textbf{Combination with Other Methods}.}
\label{table_combine_multitask}
\end{table*}

\subsection{Performance in fine-tuning domain}
\label{appendix:finetuning_domain}

Results resented in in Tab.~\ref{table3} further validate the capability preservation in domain, which illustrates SAM effectively mitigates the CF with a limited impact on fine-tuning performance in specific domains.

\subsection{Comprehensive Experimental Results Across Evaluation Datasets}
\label{appendix:A.6}

\subsubsection{Comprehensive Experimental Results on different datasets}

Based on the Llama2-7B model, our comprehensive experimental results on different datasets are presented in Tab.~\ref{table_epoch_multitask}, which presents the performance of the model across different evaluation tasks during the fine-tuning stage.

\subsubsection{Comprehensive Experimental Results across different model sizes}

On the Open-Platypus dataset, we conduct experiments across models of different sizes, with comprehensive experimental results displayed in Tab.~\ref{table_size_multitask}, which demonstrates the generality of the SAM across models of different sizes. As the comprehensive results of llama2-7B have already been presented in Tab.~\ref{table_epoch_multitask}, they will not be reiterated in Tab.~\ref{table_size_multitask}.

\subsubsection{Comprehensive Experimental Results of comparison with other advanced methods}

We conduct experiments to compare SAM with Wise-FT and Rehearsal on Llama-7B model, with comprehensive experimental results displayed in Tab.~\ref{table_compare_multitask}, which indicates SAM significantly outperforms Wise-FT and Rehearsal. Since the performance of SAM has been comprehensively demonstrated in Tab.~\ref{table_epoch_multitask}, here we only additionally present the results of Wise-FT(w/o) and Rehearsal(w/o).

\subsubsection{Comprehensive Experimental Results of complementary to existing works}
To further demonstrate the orthogonality of SAM, we combine SAM with existing anti-forgetting methods. The results, as shown in Tab.~\ref{table_combine_multitask}, indicate that our method provides incremental benefits to mitigate CF. Similarly, we only present the results of Wise-FT(w/) and Rehearsal(w/) combined with SAM.

\end{document}